\documentclass{article} 
\usepackage{iclr2015,times}
\usepackage{hyperref}
\usepackage{url}
\usepackage{amsmath}
\usepackage{amssymb}
\usepackage[ruled]{algorithm}
\usepackage{algpseudocode}
\usepackage{xspace}
\usepackage{caption}
\usepackage{subcaption}
\usepackage{multirow}
\usepackage{graphicx}

\newcommand{\randn}{\operatorname{randn}}
\newcommand{\orth}{\operatorname{orthogonalize}}
\newcommand{\eig}{\operatorname{eig}}

\newcommand{\R}{\mathbb{R}}

\DeclareMathOperator*{\argmin}{arg\,min}

\newcommand{\RandEmbed}{Rembrandt\xspace}

\iclrfinalcopy 

\begin{document}

\author{
Paul Mineiro \& Nikos Karampatziakis \\
Microsoft Cloud Information Services Lab \\
\texttt{\{pmineiro,nikosk\}@microsoft.com} \\
}

\title{Fast Label Embeddings for Extremely Large Output Spaces}

\maketitle

\begin{abstract}
Many modern multiclass and multilabel problems are characterized by
increasingly large output spaces.   For these problems, label embeddings
have been shown to be a useful primitive that can improve computational
and statistical efficiency.  In this work we utilize a correspondence
between rank constrained estimation and low dimensional label embeddings
that uncovers a fast label embedding algorithm which works in both the
multiclass and multilabel settings.  The result is a randomized algorithm
for partial least squares, whose running time is exponentially faster
than naive algorithms.  We demonstrate our techniques on two
large-scale public datasets, from the Large Scale Hierarchical Text
Challenge and the Open Directory Project, where we obtain state of the
art results.
\end{abstract}

\section{Contributions}

We provide a statistical motivation for label embedding by demonstrating
that the optimal rank-constrained least squares estimator can be
constructed from an optimal unconstrained estimator of an embedding of
the labels.  In other words, embedding can provide beneficial sample
complexity reduction even if computational constraints are not binding.

We identify a natural object to define label similarity: the expected
outer product of the conditional label probabilities.  In particular,
in conjunction with a low-rank constraint, this indicates two label
embeddings are similar when their conditional probabilities are linearly
dependent across the dataset. This unifies prior work utilizing the
confusion matrix for multiclass~\citep{bengio2010label} and the empirical
label covariance for multilabel~\citep{tai2012multilabel}.

We apply techniques from randomized linear
algebra~\citep{halko2011finding} to develop an efficient and scalable
algorithm for constructing the embeddings, essentially via a novel
randomized algorithm for partial least squares~\citep{geladi1986partial}.
Intuitively, this technique implicitly decomposes the prediction matrix
of a model which would be prohibitively expensive to form explicitly. 

\section{Description}

\subsection{Notation}

We denote vectors by lowercase letters $x$, $y$ etc.\ and matrices by
uppercase letters $W$, $Z$ etc.  The input dimension is denoted by $d$,
the output dimension by $c$ and the embedding dimension by $k$.
For an $m\times n$ matrix $X \in
\R^{m \times n}$ we use $||X||_F$ for its Frobenius norm, $X^\dagger$
for the pseudoinverse, $\Pi_{X,L}$ for the projection onto the left
singular subspace of $X$.

\begin{algorithm}[t]
  \caption{Rembrandt: Response EMBedding via RANDomized Techniques}
  \begin{algorithmic}[1]
    \Statex
    \Function{Rembrandt}{$k, X \in \R^{n \times d}, Y \in \R^{n \times c}$}
      \State $(p, q) \leftarrow (20, 1)$ \Comment{These hyperparameters rarely need adjustment.}
      \State $Q \leftarrow \randn(c, k+p)$  \label{lin:rangefindstart}
      \For{$i \in \{ 1, \ldots, q \}$} \Comment{Randomized range finder for $Y^\top \Pi_{X,L} Y$}
        \State $Z \leftarrow \argmin \| YQ - XZ \|^2_F$ \label{lin:learningstep}
        \State $Q \leftarrow \orth(Y^\top X Z)$
      \EndFor \label{lin:rangefindend} \Comment{NB: total of $(q+1)$ data passes, including next line}
      \State $F \leftarrow (Y^\top X Q)^\top (Y^\top X Q)$ \Comment{$F \in \R^{(k+p)\times(k+p)}$ is ``small''}
      \State $(V, \Sigma^2) \leftarrow \eig(F, k)$
      \State $V \leftarrow Q V$ \Comment{$V \in \R^{c \times k}$ is the embedding}
      \State \Return $(V, \Sigma)$
    \EndFunction
  \end{algorithmic}
  \label{alg:rembed}
\end{algorithm}

\subsection{Proposed Algorithm}
Our proposal is \RandEmbed, described in Algorithm \ref{alg:rembed}.
We use the top right singular
space of $\Pi_{X,L} Y$ as a label embedding, or equivalently, the top
principal components of $Y^\top \Pi_{X,L} Y$ (leveraging the fact that
the projection is idempotent).  Using randomized techniques, we can
decompose this matrix without explicitly forming it, because we can
compute the product of $\Pi_{X,L} Y$ with another matrix $Q$ via
$Y^\top \Pi_{X,L} Y Q = Y^\top X Z^*$ where 
$Z^* = \argmin_{Z \in \R^{d \times (k+p)}} \| Y Q - X Z \|^2_F$.
Algorithm \ref{alg:rembed} is a specialization of randomized PCA to this
particular form of the matrix multiplication operator.  Fortunately, because 
$X^\dagger (\Pi_{X,L} Y)_k$ is the optimal rank-constrained least squares
weight matrix, \RandEmbed is a randomized algorithm for 
partial least squares~\citep{geladi1986partial}.

Algorithm \ref{alg:rembed} is inexpensive to compute.  The matrix vector
product $Y Q$ is a sparse matrix-vector product so complexity $O(n
s k)$ depends only on the average (label) sparsity per example $s$
and the embedding dimension $k$, and is independent of the number of
classes $c$.  The fit is done in the embedding space and therefore is
independent of the number of classes $c$, and the outer product with the
predicted embedding is again a sparse product with complexity $O(n s k)$.
The orthogonalization step is $O(ck^2)$, but this is amortized over the
data set and essentially irrelevant as long as $n > c$.  Furthermore random
projection theory suggests $k$ should grow only logarithmically with $c$.

\subsection{Experiments}

\begin{table}[h]
\centering
\caption{Data sets used for experimentation and times to compute an embedding.  Timings are for a Matlab implementation on a standard desktop (dual 3.2Ghz Xeon E5-1650 CPU and 48Gb of RAM).}
\begin{tabular}{|c|c|c|c|c|c|c|c|} \hline
Dataset & Type & Modality & Examples & Features & Classes & \multicolumn{2}{c|}{\RandEmbed} \\ 
& & & & & & $k$ & Time (sec) \\ \hline
ODP & Multiclass & Text & $\sim$ 1.5M & $\sim$ 0.5M & $\sim$ 100K & 300 & 6,530 \\
LSHTC & Multilabel & Text & $\sim$ 2.4M & $\sim$ 1.6M & $\sim$ 325K & 500 & 8,006 \\ \hline
\end{tabular}
\label{tab:datasets}
\end{table}

\begin{table}[h]
\centering
\caption{ODP results.  $k=300$ for all embedding strategies. RE = \RandEmbed; CS = compressed sensing; PCA = unsupervised (feature) embedding; LT = LomTree \citep{choromanska2014logarithmic}; ``A+LR'' = logistic regression on representation A.}  
\begin{tabular}{|c|c|c|c|c|} \hline
Method & RE + LR & CS + LR & PCA + LR & LT \\ \hline
Test Error & 83.15\% & 85.14\% & 90.37\% & 93.46\% \\ \hline
\end{tabular}
\label{tab:odp}
\end{table}

\begin{table}[!h]
\centering
\caption{LSHTC results. FastXML and LPSR-NB are from \cite{prabhu2014fastxml}.  ``A+ILR'' = independent logistic regression on representation A.}
\begin{tabular}{|c|c|c|c|c|} \hline
Method &  RE ($k=800$) + ILR & RE ($k=500$) + ILR &  FastXML & LPSR-NB \\ \hline
Precision-at-1& 53.39\% & 52.84\% & 49.78\% & 27.91\% \\ \hline 
\end{tabular}
\label{tab:lshtc}
\end{table}

\bibliography{rembed.abstract}

\begin{thebibliography}{6}
\providecommand{\natexlab}[1]{#1}
\providecommand{\url}[1]{\texttt{#1}}
\expandafter\ifx\csname urlstyle\endcsname\relax
  \providecommand{\doi}[1]{doi: #1}\else
  \providecommand{\doi}{doi: \begingroup \urlstyle{rm}\Url}\fi

\bibitem[Bengio et~al.(2010)Bengio, Weston, and Grangier]{bengio2010label}
Bengio, Samy, Weston, Jason, and Grangier, David.
\newblock Label embedding trees for large multi-class tasks.
\newblock In \emph{Advances in Neural Information Processing Systems}, pp.\
  163--171, 2010.

\bibitem[Choromanska \& Langford(2014)Choromanska and
  Langford]{choromanska2014logarithmic}
Choromanska, Anna and Langford, John.
\newblock Logarithmic time online multiclass prediction.
\newblock \emph{arXiv preprint arXiv:1406.1822}, 2014.

\bibitem[Geladi \& Kowalski(1986)Geladi and Kowalski]{geladi1986partial}
Geladi, Paul and Kowalski, Bruce~R.
\newblock Partial least-squares regression: a tutorial.
\newblock \emph{Analytica chimica acta}, 185:\penalty0 1--17, 1986.

\bibitem[Halko et~al.(2011)Halko, Martinsson, and Tropp]{halko2011finding}
Halko, Nathan, Martinsson, Per-Gunnar, and Tropp, Joel~A.
\newblock Finding structure with randomness: Probabilistic algorithms for
  constructing approximate matrix decompositions.
\newblock \emph{SIAM review}, 53\penalty0 (2):\penalty0 217--288, 2011.

\bibitem[Prabhu \& Varma(2014)Prabhu and Varma]{prabhu2014fastxml}
Prabhu, Yashoteja and Varma, Manik.
\newblock Fastxml: a fast, accurate and stable tree-classifier for extreme
  multi-label learning.
\newblock In \emph{Proceedings of the 20th ACM SIGKDD international conference
  on Knowledge discovery and data mining}, pp.\  263--272. ACM, 2014.

\bibitem[Tai \& Lin(2012)Tai and Lin]{tai2012multilabel}
Tai, Farbound and Lin, Hsuan-Tien.
\newblock Multilabel classification with principal label space transformation.
\newblock \emph{Neural Computation}, 24\penalty0 (9):\penalty0 2508--2542,
  2012.

\end{thebibliography}
\bibliographystyle{iclr2015}

\end{document}